\documentclass[11,letterpaper]{article}

\usepackage[margin=1.0in]{geometry}

\usepackage{sectsty}

\usepackage{notoccite}
\usepackage{graphics} 
\usepackage{amsmath} 
\usepackage{amssymb}  
\usepackage{color,xspace}
\usepackage{graphicx}
\usepackage{subcaption}
\usepackage{cite}
\usepackage{cleveref}
\usepackage{bm}
\usepackage{bbm}
\usepackage{tikz}
\usepackage{epstopdf} 
\usepackage{ntheorem}
\usepackage{enumitem}
\usepackage{mdframed}   
\usepackage[normalem]{ulem}

\usepackage{xcolor,cancel}

\usepackage[many]{tcolorbox}
\newtcolorbox{cross}{blank,breakable,parbox=false,
	overlay={\draw[red,line width=5pt] (interior.south west)--(interior.north east);
		\draw[red,line width=5pt] (interior.north west)--(interior.south east);}}

\newcommand{\specialcell}[2][c]{%
	\begin{tabular}[#1]{@{}c@{}}#2\end{tabular}}

\usepackage{caption} 
\newcommand{\new}[1]{{\color{black}{#1}}}

\newtheorem{theorem}{Theorem}
\newtheorem{lemma}{Lemma}    

\newtheorem{remark}{Remark}

\newtheorem{proposition}{Proposition}    
    
\newtheorem{definition}{Definition}

\newcommand{\R}{\mathbb{R}}

\newcommand{\bx}{\mathbf{x}}
\newcommand{\by}{\mathbf{y}}

\newcommand{\norm}[1]{\left|\left| #1 \right|\right|}

\title{{\LARGE \bf Efficient and Accurate Estimation of Lipschitz Constants for Deep Neural Networks}}
\author{Mahyar Fazlyab \quad Alexander Robey \quad Hamed Hassani  \\ Manfred Morari \quad  George J. Pappas
	\thanks{The authors are with the Department of Electrical and Systems Engineering, University of Pennsylvania. Email: \{mahyarfa,arobey1,hassani,morari,gpappas\}@seas.upenn.edu.}} 

\date{}

\begin{document}
	\pagestyle{plain}
	\maketitle

	\begin{abstract}
		
		Tight estimation of the Lipschitz constant for deep neural networks (DNNs) is useful in many applications ranging from robustness certification of classifiers to stability analysis of closed-loop systems with reinforcement learning controllers. Existing methods in the literature for estimating the Lipschitz constant suffer from either lack of accuracy or poor scalability. In this paper, we present a convex optimization framework to compute guaranteed upper bounds on the Lipschitz constant of DNNs both accurately and efficiently. Our main idea is to interpret activation functions as gradients of convex potential functions. Hence, they satisfy certain properties that can be described by quadratic constraints. This particular description allows us to pose the Lipschitz constant estimation problem as a semidefinite program (SDP). The resulting SDP can be adapted to increase either the estimation accuracy (by capturing the interaction between activation functions of different layers) or scalability (by decomposition and parallel implementation).  We illustrate the utility of our approach with a variety of experiments on randomly generated networks and on classifiers trained on the MNIST and Iris datasets. In particular, we experimentally demonstrate that our Lipschitz bounds are the most accurate compared to those in the literature. We also study the impact of adversarial training methods on the Lipschitz bounds of the resulting classifiers and show that our bounds can be used to efficiently provide robustness guarantees.

	\end{abstract}
	
	\section{Introduction}
	
	A function $f \colon \mathbb{R}^n \to \mathbb{R}^m$ is globally Lipschitz continuous on $\mathcal{X} \subseteq \mathbb{R}^n$ if there exists a nonnegative constant $L \geq 0$ such that
	\begin{align} \label{eq: Lipschitz continuity}
		\|f(x)-f(y)\| \leq L \|x-y\| \ \text{ for all } x,y \in \mathcal{X}.
	\end{align}
	The smallest such $L$ is called \emph{the} Lipschitz constant of $f$. The Lipschitz constant is the maximum ratio between variations in the output space and variations in the input space of $f$ and thus is a measure of sensitivity of the function with respect to input perturbations.
	
	When a function $f$ is characterized by a deep neural network (DNN), tight bounds on its Lipschitz constant can be extremely useful in a variety of applications. In classification tasks, for instance, $L$ can be used as a certificate of robustness of a neural network classifier to adversarial attacks if it is estimated tightly \cite{szegedy2013intriguing}. In deep reinforcement learning, tight bounds on the Lipschitz constant of a DNN-based controller can be directly used to analyze the stability of the closed-loop system. Lipschitz regularity can also play a key role in derivation of generalization bounds \cite{bartlett2017spectrally}. In these applications and many others, it is essential to have tight bounds on the Lipschitz constant of DNNs. However, as DNNs have highly complex and non-linear structures, estimating the Lipschitz constant both accurately and efficiently has remained a significant challenge. 
	

	\medskip \noindent \textbf{Our contributions.} In this paper we propose a novel convex programming framework to derive~tight bounds on the global Lipschitz constant of deep feed-forward neural networks. Our framework yields significantly more \emph{accurate} bounds compared to the state-of-the-art and lends itself to a distributed implementation, leading to \emph{efficient} computation of the bounds for large-scale networks. 
	
	\medskip  \noindent \textbf{Our approach.} We use the fact that nonlinear activation functions used in neural networks are gradients of convex functions; hence, as operators, they satisfy certain properties that can be abstracted as quadratic constraints on their input-output values. This particular abstraction allows us to pose the Lipschitz estimation problem as a semidefinite program (SDP), which we call \texttt{LipSDP}. A striking feature of \texttt{LipSDP} is its flexibility to span the trade-off between estimation accuracy and computational efficiency by adding or removing extra decision variables. In particular, for a neural network with $\ell$ layers and a total of $n$ hidden neurons, the number of decision variables can vary from $\ell$ (least accurate but most scalable) to $O(n)$ (most accurate but least scalable).  As such, we derive several distinct yet related formulations of \texttt{LipSDP} that span this trade-off. To scale each variant of \texttt{LipSDP} to larger networks, we also propose a distributed implementation. 
	
	\medskip  \noindent \textbf{Our results.} We illustrate our approach in a variety of experiments on both randomly generated networks as well as networks trained on the MNIST \cite{lecun1998mnist} and Iris \cite{Dua:2019} datasets. First, we show empirically that our Lipschitz bounds are the most accurate compared to all other existing methods of which we are aware. In particular, our experiments on neural networks trained for MNIST show that our bounds almost coincide with the \emph{true} Lipschitz constant and outperform all comparable methods.  For details, see Figure \ref{fig:comp-mnist}.  Furthermore, we investigate the effect of two robust training procedures \cite{madry2017towards,kolter2017provable} on the Lipschitz constant for networks trained on the MNIST dataset.  Our results suggest that robust training procedures significantly decrease the Lipschitz constant of the resulting classifiers.  Moreover, we use the Lipschitz bound for two robust training procedures to derive non-vacuous lower bounds on the minimum adversarial perturbation necessary to change the classification of any instance from the test set. For details, see Figure \ref{fig:robust-training}.

	\medskip  \noindent \textbf{Related work.} The problem of estimating the Lipschitz constant for neural networks has been studied in several works. In \cite{szegedy2013intriguing}, the authors estimate the global Lipschitz constant of DNNs by the product of Lipschitz constants of individual layers. This approach is scalable and general but yields trivial bounds. We are only aware of two other methods that give non-trivial upper bounds on the global Lipschitz constant of fully-connected neural networks and can scale to networks with more than two hidden layers.  In \cite{combettes_lipschitz}, Combettes and Pesquet derive bounds on Lipschitz constants by treating the activation functions as non-expansive averaged operators. The resulting algorithm scales well with the number of hidden units per layer, but very poorly (in fact exponential) with the number of layers. In \cite{virmaux2018lipschitz}, Virmaux and Scaman decompose the weight matrices of a neural network via singular value decomposition and approximately solve a convex maximization problem over the unit cube.  Notably, estimating the Lipschitz constant using the method in \cite{virmaux2018lipschitz} is intractable even for small networks; indeed, the authors of \cite{virmaux2018lipschitz} use a greedy algorithm to compute a bound, which may underapproximate the Lipschitz constant. Bounding Lipschitz constants for the specific case of convolutional neural networks (CNNs) has also been addressed in \cite{balan2017lipschitz,zou2018lipschitz,bartlett2017spectrally}. 
	
	Using Lipschitz bounds in the context of adversarial robustness and safety verification has also been addressed in several works \cite{weng2018evaluating,ruan2018reachability,weng2018towards}. In particular, in \cite{weng2018evaluating}, the authors convert the robustness analysis problem into a local Lipschitz constant estimation problem, where they estimate this local constant by a set of independently and identically sampled local gradients. This algorithm is scalable but is not guaranteed to provide upper bounds. In a similar work, the authors of \cite{weng2018towards} exploit the piece-wise linear structure of ReLU functions to estimate the local Lipschitz constant of neural networks. In \cite{fazlyab2019safety}, the authors use quadratic constraints and semidefinite programming to analyze local (point-wise) robustness of neural networks. In contrast, our Lipschitz bounds can be used as a global certificate of robustness and are agnostic to the choice of the test data.

	\subsection{Motivating applications}
	
	%
	We now enumerate several applications that highlight the importance of estimating the Lipschitz constant of DNNs accurately and efficiently.
	
	\medskip  \noindent \textbf{Robustness certification of classifiers.} In response to fragility of DNNs to adversarial attacks, there has been considerable effort in recent years to improve the robustness of neural networks against adversarial attacks and input perturbations \cite{goodfellow6572explaining,papernot2016distillation,zheng2016improving,kurakin2016adversarial,madry2017towards,kolter2017provable}. In order to certify and/or improve the robustness of neural networks, one must be able to bound the possible outputs of the neural network over a region of input space. This can be done either locally around a specific input \cite{bastani2016measuring,tjeng2017evaluating,gehr2018ai2,singh2018fast,dutta2018output,raghunathan2018certified,raghunathan2018semidefinite,fazlyab2019safety,kolter2017provable,jordan2019provable,wong2018scaling,zhang2018efficient}, or globally by bounding the sensitivity of the function to input perturbations, i.e., the Lipschitz constant \cite{huster2018limitations,szegedy2013intriguing,qian2018l2,weng2018evaluating}.   Indeed, tight upper bounds on the Lipschitz constant can be used to derive non-vacuous lower bounds on the magnitudes of perturbations necessary to change the decision of neural networks. Finally, an efficient computation of these bounds can be useful in either assessing robustness after training \cite{raghunathan2018certified,raghunathan2018semidefinite,fazlyab2019safety} or promoting robustness during training \cite{kolter2017provable,tsuzuku2018lipschitz}. In the experiments section, we explore this application in depth.

	\medskip  \noindent \textbf{Stability analysis of closed-loop systems with learning controllers.} 
	A central problem in learning-based control is to provide stability or safety guarantees for a feedback control loop when a learning-enabled component, such as a deep neural network, is introduced in the loop \cite{aswani2013provably,berkenkamp2017safe,jin2018stability}. The Lipschitz constant of a neural network controller bounds its gain. Therefore a tight estimate  can be useful for certifying the stability of the closed-loop system.

	\ifx
	\section{Lipschitz constant as a global certificate of robustness}
	Despite their success in various domains, deep neural networks (DNNs) are vulnerable to adversarial attacks \cite{szegedy2013intriguing}. More specifically, carefully designed undetectable perturbations in the input can drastically change the output of the neural network. These perturbations can be generated, for example, by maximizing the prediction error over a neighborhood of an input data point (adversarial attacks). In practical implementations, they can also be non-adversarial and can occur due to compression, resizing, and cropping. In this context, we say that a neural network is locally robust (or point-wise robust) if the neural network's prediction remains the same in a bounded neighborhood of a query data point.
	
	To certify the robustness of neural networks to adversarial attacks and input perturbations, one must be able to bound the possible outputs of the neural network over a region of input space. This can be done either locally around a specific input \cite{bastani2016measuring,tjeng2017evaluating,dutta2018output,raghunathan2018certified,raghunathan2018semidefinite,fazlyab2019safety,kolter2017provable,jordan2019provable,wong2018scaling,zhang2018efficient}, or globally by bounding the sensitivity of the function to input perturbations, i.e., the Lipschitz constant \cite{huster2018limitations,szegedy2013intriguing,qian2018l2,weng2018evaluating}. 
	%
	%
	Since, computing the Lipschitz constant is an NP-hard problem \cite{virmaux2018lipschitz}, one has to settle for finding non-trivial upper bounds. Indeed, having a sharp bound on the Lipschitz constant is important in that large bounds do not automatically translate into existence of adversarial examples \textcolor{red}{kind of awkward here}; however, small bounds guarantee that no such examples can appear \cite{szegedy2013intriguing}. Indeed, tight upper bounds on the Lipschitz it can be used to derive non-vacuous lower bounds on the magnitudes of perturbations necessary to change the decision of neural networks (see Proposition \ref{prop: relation between L and eps} and Figure \ref{fig: robustness_certification}). 
	
	Consider a neural-network classifier $f\colon \mathbb{R}^n \to \mathbb{R}^k$ with Lipschitz constant $L_f$ in the $\ell_2$-norm. Let $\epsilon>0$ be given and consider the inequality
	\begin{align} \label{eq: local robustness inequality}
		L_f \leq \frac{1}{\epsilon \sqrt{2}} \min_{1 \leq i \neq j \leq k} \ |f_i(x^\star)-f_j(x^\star)|.
	\end{align}
	Then \eqref{eq: local robustness inequality} implies $C(x):=\mathrm{argmax}_{1 \leq i \leq k} f_i(x) = C(x^\star)$ for all $x\in\mathcal{A}(x^{\star})$.
	

The inequality in \eqref{eq: local robustness inequality} provides us with a simple and computationally efficient test for assessing the point-wise robustness of a neural network. According to \eqref{eq: local robustness inequality}, a more accurate estimation of the Lipschitz constant directly increases the maximum perturbation $\epsilon$ that can be certified for each test example. This makes the framework suitable for model selection, wherein one wishes to select the model that is most robust to adversarial perturbations from a family of proposed classifiers \cite{peck2017lower}.
\fi


\medskip  \noindent \textbf{Notation.} We denote the set of real $n$-dimensional vectors by $\mathbb{R}^n$, the set of $m\times n$-dimensional matrices by $\mathbb{R}^{m\times n}$, and the $n$-dimensional identity matrix by $I_n$. We denote by $\mathbb{S}^{n}$, $\mathbb{S}_{+}^n$, and $\mathbb{S}_{++}^n$ the sets of $n$-by-$n$ symmetric, positive semidefinite, and positive definite matrices, respectively. The $p$-norm ($p \geq 1$) is denoted by $\|\cdot\|_p \colon \mathbb{R}^n \to \mathbb{R}_{+}$.  The $\ell_2$-norm of a matrix $W\in \R^{m\times n}$ is the largest singular value of $W$.  We denote the $i$-th unit vector in $\mathbb{R}^n$ by $e_i$. We write $\mathrm{diag}(a_1, ..., a_n)$ for a diagonal matrix whose diagonal entries starting in the upper left corner are $a_1, \cdots, a_n$.

\section{LipSDP: Lipschitz certificates via semidefinite programming}

\subsection{Problem statement} 
\label{sect:prob-statement}
Consider an $\ell$-layer feed-forward neural network $f(x) \colon \mathbb{R}^{n_0} \to \mathbb{R}^{n_{\ell+1}}$ described by the following recursive equations:
\begin{align} \label{eq: DNN model 0}
	x^0 =x, \quad x^{k+1} =\phi(W^k x^k + b^k) \ \text{ for } k=0,\cdots,\ell-1, \quad f(x) = W^\ell x^\ell + b^{\ell}.
\end{align}
Here $x \in \mathbb{R}^{n_0}$ is an input to the network and $W^{k} \in \mathbb{R}^{n_{k+1}\times n_k}$ and $b^k \in \mathbb{R}^{n_{k+1}}$ are the weight matrix and bias vector for the $k$-th layer. The function $\phi$ is the concatenation of activation functions at each layer, i.e., it is of the form $\phi(x) = [\varphi({x}_1) \ \cdots \ \varphi({x}_{n})]^\top$.
%
In this paper, our goal is to find tight bounds on the Lipschitz constant of the map $x \mapsto f(x)$ in $\ell_2$-norm.  More precisely, we wish to find the smallest constant $L_2\geq 0$ such that $\|f(x)-f(y)\|_2 \leq L_2 \|x-y\|_2 \ \text{for all } x,y \in \mathbb{R}^{n_0}$.

The main source of difficulty in solving this problem is the presence of the nonlinear activation functions. To combat this difficulty, our main idea is to abstract these activation functions by a set of constraints that they impose on their input and output values. Then any property (including Lipschitz continuity) that is satisfied by our abstraction will also be satisfied by the original network.

\subsection{Description of activation functions by quadratic constraints}

In this section, we introduce several definitions and lemmas that characterize our abstraction of nonlinear activation functions.  These results are crucial to the formulation of an SDP that can bound the Lipschitz constants of  networks in Section~\ref{sect:main-result}.

\begin{figure}
	\centering
	\includegraphics[width=1\linewidth]{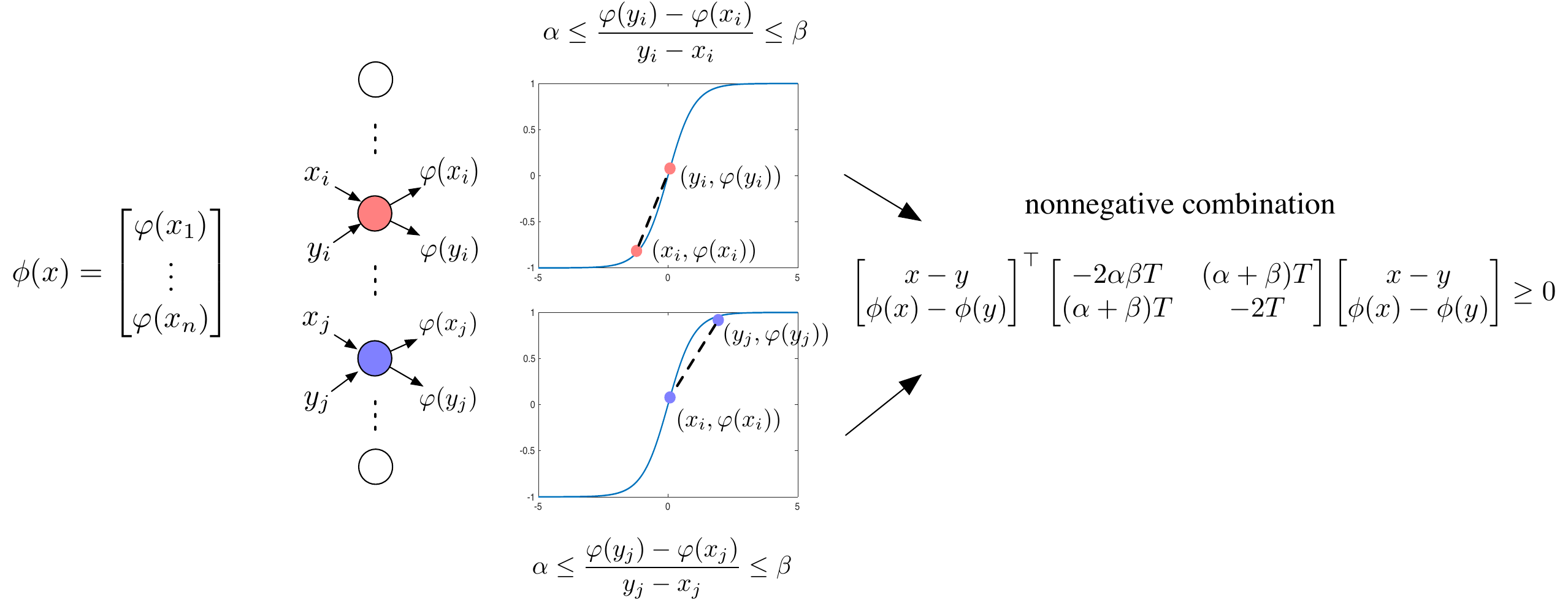}
	\caption{An illustrative description of encoding activation functions by quadratic constraints.}
	\label{fig: repeated_nonlinearities}
\end{figure}

\begin{definition}[Slope-restricted non-linearity] \label{def: slope restricted nonlinearity}
	A function $\varphi \colon \mathbb{R} \to \mathbb{R}$ is slope-restricted on $[\alpha,\beta]$ where $0\leq \alpha < \beta <\infty$ if
	\begin{align} \label{eq: activation function slope condition}
		\alpha \leq \frac{\varphi(y)-\varphi(x)}{y-x} \leq \beta \quad \forall x,y \in \mathbb{R}.
	\end{align}
\end{definition}
The inequality in \eqref{eq: activation function slope condition} simply states that the slope of the chord connecting any two points on the curve of the function $x \mapsto \varphi(x)$ is at least $\alpha$ and at most $\beta$ (see Figure \ref{fig: repeated_nonlinearities}). By multiplying all sides of \eqref{eq: activation function slope condition} by $(y-x)^2$, we can write the slope restriction condition as $\alpha (y-x)^2 \leq (\varphi(y)-\varphi(x))(y-x) \leq \beta (y-x)^2$.
%
%
By the left inequality, the operator $\varphi(x)$ is strongly monotone with parameter $\alpha$ \cite{ryu2016primer}, or equivalently the anti-derivative function $\int \varphi(x)dx$ is strongly convex with parameter $\alpha$. By the right-hand side inequality, $\varphi(x)$ is one-sided Lipschitz with parameter $\beta$. Altogether, the preceding inequalities state that the anti-derivative function $\int \varphi(x)dx$ is $\alpha$-strongly convex and $\beta$-smooth. 

Note that all activation functions used in deep learning satisfy the slope restriction condition in \eqref{eq: activation function slope condition} for some $0 \leq \alpha < \beta <\infty$. For instance, the ReLU, tanh, and sigmoid activation functions are all slope restricted with $\alpha = 0$ and $\beta=1$. More details can be found in \cite{fazlyab2019safety}.  

\begin{definition}[Incremental Quadratic Constraint \cite{accikmecse2011observers}] \label{def: Incremental}\normalfont A function $\phi \colon \mathbb{R}^n \to \mathbb{R}^n$ satisfies the incremental quadratic constraint defined by $\mathcal{Q} \subset \mathbb{S}^{2n}$ if for any $Q \in \mathcal{Q}$ and  $x,y \in \mathbb{R}^n$,
	\begin{align}
		\begin{bmatrix}
			x-y \\ \phi(x)-\phi(y)
		\end{bmatrix}^\top Q \begin{bmatrix}
			x-y \\ \phi(x)-\phi(y)
		\end{bmatrix} \geq 0.
	\end{align}	
\end{definition}

In the above definition, $\mathcal{Q}$ is the set of all \emph{multiplier matrices} that characterize $\phi$, and is a convex cone by definition. As an example, the softmax operator $\phi(x)=(\sum_{i=1}^{n} \exp(x_i))^{-1}[\exp(x_1) \cdots \exp(x_n)]^\top$ is the gradient of the convex function $\psi(x)=\log(\sum_{i=1}^{n} \exp(x_i))$. This function is smooth and strongly convex with paramters $\alpha=0$ and $\beta=1$ \cite{boyd2004convex}. For this class of functions, it is known that the gradient function $\phi(x)=\nabla \psi(x)$ satisfies the quadratic inequality \cite{nesterov2013introductory}
\begin{align}
	\begin{bmatrix}
		x-y \\ \phi(x)-\phi(y)
	\end{bmatrix}^\top \begin{bmatrix}
		-2\alpha \beta I_n & (\alpha+\beta)I_n \\ (\alpha+\beta) I_n& -2I_n
	\end{bmatrix}  \begin{bmatrix}
		x-y \\ \phi(x)-\phi(y)
	\end{bmatrix} \geq 0.
\end{align}
Therefore, the softmax operator satisfies the incremental quadratic constraint defined by $\mathcal{Q}=\{\lambda M \mid \lambda \geq 0\}$, where $M$ the middle matrix in the above inequality.

To see the connection between incremental quadratic constraints and slope-restricted nonlinearities, note that \eqref{eq: activation function slope condition} can be equivalently written as the single inequality
\begin{align} \label{def: slope restricted nonlinearity 11}
	(\frac{\varphi(y)-\varphi(x)}{y-x}-\alpha)(\frac{\varphi(y)-\varphi(x)}{y-x}-\beta) \leq 0.
\end{align}
Multiplying through by $(y-x)^2$ and rearranging terms, we can write \eqref{def: slope restricted nonlinearity 11} as
\begin{align} \label{eq: slope restricted qc}
	\begin{bmatrix}
		x-y \\ \varphi(x)-\varphi(y)
	\end{bmatrix}^\top \begin{bmatrix}
		-2\alpha \beta  & \alpha+\beta \\ \alpha+\beta & -2
	\end{bmatrix} \begin{bmatrix}
		x-y \\ \varphi(x)-\varphi(y)
	\end{bmatrix} \geq 0,
\end{align}
which, in view of Definition \ref{def: Incremental}, is an incremental quadratic constraint for $\varphi$. From this perspective, incremental quadratic constraints generalize the notion of slope-restricted nonlinearities to multi-variable vector-valued nonlinearities. \medskip



%
\begin{lemma} \label{lem: activation function slope condition vector}
	Suppose $\varphi \colon \mathbb{R} \to \mathbb{R}$ is slope-restricted on $[\alpha,\beta]$. Define the set
	\begin{align} \label{lem: activation function slope condition vector 1}
		\mathcal{T}_n = \{T \in \mathbb{S}^n \mid T = \sum_{i=1}^{n} \lambda_{ii} e_i e_i^\top,  \new{\lambda_{ii}\geq 0} \}. 
	\end{align}
	Then for any $T \in \mathcal{T}_n$ the vector-valued function $\phi(x) = [\varphi(x_1) \cdots \varphi(x_n)]^\top \colon \mathbb{R}^n \to \mathbb{R}^n$ satisfies 
	\begin{align} \label{eq: activation function slope condition vector}
		\begin{bmatrix}
			x-y \\ \phi(x)-\phi(y)
		\end{bmatrix}^\top \begin{bmatrix}
			-2\alpha \beta T & (\alpha+\beta)T \\ (\alpha+\beta)T & -2T
		\end{bmatrix}  \begin{bmatrix}
			x-y \\ \phi(x)-\phi(y)
		\end{bmatrix} \geq 0 \ \text{ for all } x,y \in \mathbb{R}^n.
	\end{align}
\end{lemma}
%

We will see in the next section that the matrix $T$ that parameterizes the multiplier matrix in \eqref{eq: activation function slope condition vector} appears as a decision variable in an SDP, in which the objective is to find an admissible $T$ that yields the tightest bound on the Lipschitz constant. 

\medskip

\subsection{LipSDP for single-layer neural network}
\label{sect:main-result}
To develop an optimization problem to estimate the Lipschitz constant of a fully-connected feed-forward neural network, the key insight is that the Lipschitz condition in \eqref{eq: Lipschitz continuity} is in fact equivalent to an incremental quadratic constraint for the map $x \mapsto f(x)$ characterized by the neural network. By coupling this to the incremental quadratic constraints satisfied by the cascade combination of the activation functions \cite{fazlyab2018analysis}, we can develop an SDP to minimize an upper bound on the Lipschitz constant of $f$. This result is formally stated in the following theorem.
%
%
\begin{theorem}[Lipshitz certificates for single-layer neural networks] \label{thm: Lipschitz one layer}
	Consider a single-layer neural network described by $f(x) = W^1 \phi(W^0 x + b^0) + b^1$.
	%
	%
	Suppose $\phi(x) \colon \mathbb{R}^{n} \to \mathbb{R}^{n} = [\varphi(x_1) \cdots \varphi(x_{n})]$, where $\varphi$ is slope-restricted in the sector $[\alpha,\beta]$. Define $\mathcal{T}_{n}$ as in \eqref{lem: activation function slope condition vector 1}. Suppose there exists a $\rho>0$ such that the matrix inequality
	\begin{align} \label{thm: Lipschitz one layer 2}
		M(\rho,T) := \begin{bmatrix}
			-2\alpha \beta {W^0}^\top T W^0 - \rho I_{n_0} & (\alpha+\beta) {W^0}^\top T  \\ (\alpha+\beta) TW^0 & -2T+{W^1}^\top W^1
		\end{bmatrix}\preceq 0,
	\end{align}
	holds for some $T \in \mathcal{T}_{n}$. Then $\|f(x)-f(y)\|_2 \leq \sqrt{\rho} \|x-y\|_2$ for all  $x,y \in \mathbb{R}^{n_0}$.
	%
\end{theorem}

Theorem \ref{thm: Lipschitz one layer} provides us with a sufficient condition for $L_2=\sqrt{\rho}$ to be an upper bound on the Lipschitz constant of $f(x)=W^1 \phi(W^0 x + b^0) + b^1$. In particular, we can find the tightest bound by solving the following optimization problem:
\begin{align} \label{eq: SDP one layer}
	\mathrm{minimize} \quad \rho \quad \text{ subject to} \quad M(\rho,T)\preceq 0 \quad \text{and} \quad T \in \mathcal{T}_{n},
\end{align}
where the decision variables are $(\rho,T) \in \mathbb{R}_{+} \times \mathcal{T}_{n}$. Note that $M(\rho,T)$ is linear in $\rho$ and $T$ and the set $\mathcal{T}_{n}$ is convex. Hence, \eqref{eq: SDP one layer} is an SDP, which can be solved numerically for its global minimum.
%
\subsection{LipSDP for multi-layer neural networks}
%
We now consider the multi-layer case. Assuming that all the activation functions are the same, we can write the neural network model in \eqref{eq: DNN model 0} compactly as
\begin{align}  
	B\bx = \phi(A\mathbf{x} + b) \label{eq:multi-layer-eqs} \quad \text{and} \quad f(x) = C\bx + b^{\ell},
\end{align}
where $\bx = [{x^0}^\top \ {x^1}^\top \cdots  {x^{\ell}}^\top]^\top$ is the concatenation of the input and the activation values, and the matrices $b$, $A$, $B$ and $C$ are given by \cite{fazlyab2019safety}
\begin{align} \label{eq: A and B matrices}
	A &= \begin{bmatrix}
		W^0 & 0 & \hdots & 0 & 0 \\
		0 & W^1 & \hdots & 0 & 0 \\
		\vdots & \vdots & \ddots & \vdots & \vdots \\
		0 & 0 & \hdots & W^{\ell-1} & 0
	\end{bmatrix}, \ B = \begin{bmatrix}
		0 & I_{n_1} & 0 & \hdots & 0 \\
		0 & 0 & I_{n_2} & \hdots & 0 \\
		\vdots & \vdots & \vdots & \ddots & \vdots \\
		0 & 0 & 0 & \hdots & I_{n_{\ell}}
	\end{bmatrix}, \\ \nonumber C &= \begin{bmatrix}
		0 & \hdots & 0 & W^{\ell}
	\end{bmatrix}, \ b = \begin{bmatrix}
		{b^0}^\top & \cdots & {b^{\ell-1}}^\top
	\end{bmatrix}^\top.
\end{align}
The particular representation in \eqref{eq:multi-layer-eqs} facilitates the extension of \texttt{LipSDP} to multiple layers, as stated in the following theorem.
%
\begin{theorem}[Lipschitz certificates for multi-layer neural networks]
	\label{thm:multi-layer-thm}
	Consider an $\ell$-layer fully connected neural network described by \eqref{eq: DNN model 0}.  Let $n=\sum_{k=1}^{\ell} n_k$ be the total number of hidden neurons and suppose the activation functions are slope-restricted in the sector $[\alpha,\beta]$. Define $\mathcal{T}_n$ as in \eqref{lem: activation function slope condition vector 1}. Define $A$ and $B$ as in \eqref{eq: A and B matrices}. Consider the matrix inequality
	\begin{align} \label{thm:multi-layer-thm 2}
		M(\rho,T) = \begin{bmatrix}
			A \\ B
		\end{bmatrix}^\top \begin{bmatrix}
			-2\alpha \beta T & (\alpha+\beta)T \\ (\alpha+\beta)T & -2T
		\end{bmatrix} \begin{bmatrix}
			A \\ B
		\end{bmatrix} + \begin{bmatrix}
			-\rho I_{n_0} & 0 & \hdots & 0 \\
			0 & 0 & \hdots & 0 \\
			\vdots & \vdots & \ddots & \vdots \\
			0 & 0 & \hdots & (W^{\ell})^\top W^{\ell} 
		\end{bmatrix}\preceq 0.
	\end{align}
		If \eqref{thm:multi-layer-thm 2} is satisfied for some $(\rho,T) \in \mathbb{R}_{+} \times \mathcal{T}_n$, then 
		$\norm{f(x) - f(y)}_2 \leq \sqrt{\rho} \norm{x-y}_2$, $\forall x, y\in\R^{n_0}$.
		%
	\end{theorem}
	In a similar way to the single-layer case, we can find the best bound on the Lipschitz constant by solving the SDP in \eqref{eq: SDP one layer} with $M(\rho,T)$ defined as in \eqref{thm:multi-layer-thm 2}.
	\begin{remark}\normalfont We have only considered the $\ell_2$ norm in our exposition. By using the inequality $\|x\|_p \leq n^{\frac{1}{p}-\frac{1}{q}} \|x\|_q$, the $\ell_2$-Lipschitz bound implies
		$$
		m^{-(\frac{1}{p}-\frac{1}{2})} \|f(y)-f(x)\|_p \leq \|f(y)-f(x)\|_2 \leq  L_2 \|y-x\|_2 \leq n^{\frac{1}{2}-\frac{1}{q}} L_2 \|y-x\|_q,
		$$
		or, equivalently,
		$
		\|f(y)-f(x)\|_p \leq n^{(\frac{1}{2}-\frac{1}{q})} m^{(\frac{1}{p}-\frac{1}{2})} L_2 \|y-x\|_q.
		$
		Hence, $n^{(\frac{1}{2}-\frac{1}{q})} m^{(\frac{1}{p}-\frac{1}{2})} L_2$ is a Lipschitz constant of $f$ when $\ell_q$ and $\ell_p$ norms are used in the input and output spaces, respectively. We can also extend our framework to accommodate quadratic norms $\|x\|_P = \sqrt{x^\top P x} $, where $P \in \mathbb{S}_{++}^n$.
	\end{remark}
	
	\subsection{Variants of LipSDP: reconciling accuracy and efficiency}
	\label{sect:variants}
	%
	
	\new{In \texttt{LipSDP}, there are $n+1$ decision variables $\rho\geq 0, \ \lambda_{ii}, \ 1 \leq i \leq n$, where $n$ is the total number of hidden neurons. Using all these decision variables would provide the tightest convex relaxation in our formulation. However, solving this SDP with all the decision variables included is impractical for large networks. Nevertheless, we can consider a hierarchy of relaxations of \texttt{LipSDP} by removing a subset of the decision variables. Below, we give a brief description of the efficiency and accuracy of each variant.  Throughout, we let $n$ be the total number of neurons and $\ell$ the number of hidden layers.}
	
	%
	\begin{enumerate}
		\item \textbf{LipSDP-Neuron} has $n+1$ decision variables. 
		For this case, we have $T = \mathrm{diag}(\lambda_{11},\cdots,\lambda_{nn})$.
		\item \textbf{LipSDP-Layer} considers only one constraint per layer, resulting in $\ell+1$ decision variables. It is the most scalable and least accurate method. For this variant, we have $T = \mathrm{blkdiag}(\lambda_1 I_{n_1},\cdots,\lambda_{\ell} I_{n_{\ell}})$.
	\end{enumerate}
	%
	\textbf{Parallel implementation by splitting}. The Lipschitz constant of the composition of two or more functions can be bounded by the product of the Lispchtiz constants of the individual functions.  By splitting a neural network up into small sub-networks, one can first bound the Lipschitz constant of each sub-network and then multiply these constants together to obtain a Lipschitz constant for the entire network.  Because sub-networks do not share weights, it is possible to compute the Lipschitz constants for each sub-network in parallel. This greatly improves the scalability of of each variant of \texttt{LipSDP} with respect to the total number of activation functions in the network. 
	
	\medskip
	
	\noindent \new{\textbf{Repeated nonlinearities.} The quadratic constraint in Lemma \ref{lem: activation function slope condition vector} is valid for functions of the form $\phi(x)= [\varphi_1(x_1) \cdots \varphi_n(x_n)]^\top$ as along as all $\varphi_i$'s are slope-restricted on $[\alpha,\beta]$. Therefore, this quadratic constraint does not exploit the fact that the same activation functions are used. In Appendix~\ref{appendix: LipSDP and repeated nonlinearities} we derive more refined quadratic constraints that exploit this additional structure when one of the points in the definition of Lipschitz continuity is fixed. The resulting SDP would be tighter but more expensive to solve.}

	\section{Experiments}
	\begin{table}[!t]
		\begin{minipage}[t]{.48\linewidth}
			\centering
			\begin{tabular}{|c|c|c|c|c|} \hline
				$n$ & \specialcell{LipSDP-\\Neuron} & \specialcell{LipSDP-\\Layer} \\ \hline
				500 & 5.22 & 2.85  \\ \hline
				1000 & 27.91 & 17.88  \\ \hline
				1500 & 82.12 & 58.61  \\ \hline
				2000 & 200.88 & 146.09  \\ \hline 
				2500 & 376.07 & 245.94  \\ \hline
				3000 & 734.63 & 473.25 \\ \hline
			\end{tabular}
			\caption{Computation time in seconds for evaluating Lipschitz bounds of one-hidden-layer neural networks with a varying number of hidden units.  A plot showing the Lipschitz constant for each network tested in this table has been provided in the Appendix.}
			\label{tab:scalability}
		\end{minipage} \quad
		\begin{minipage}[t]{.48\linewidth}
			\centering
			\begin{tabular}{|c|c|c|c|} \hline
				$\ell$ & \specialcell{LipSDP- \\ Neuron} & \specialcell{LipSDP-\\Layer} \\ \hline
				5 & 20.33 & 3.41 \\ \hline
				10 & 32.18 & 7.06 \\ \hline
				50 & 87.45 & 25.88 \\ \hline
				100  & 135.85 & 40.39  \\ \hline
				200  & 221.2 & 64.90 \\ \hline
				500 & 707.56 & 216.49  \\ \hline
			\end{tabular}
			\caption{Computation time in seconds for computing Lipschitz bounds of $\ell$-hidden-layer neural networks with 100 activation functions per layer.  For \texttt{LipSDP-Neuron} and \texttt{LipSDP-Layer}, we split each network up into 5-layer sub-networks.}
		\end{minipage} 
	\end{table}
	
	In this section we describe several experiments that highlight the key aspects of this work.  In particular, we show empirically that our bounds are much tighter than any comparable method, we study the impact of robust training on our Lipschitz bounds, and we analyze the scalability of our methods.
	
	\medskip  \noindent \textbf{Experimental setup.} For our experiments we used MATLAB, the CVX toolbox \cite{grant2008cvx} and MOSEK \cite{mosek} on a 9-core CPU with 16GB of RAM to solve the SDPs.  All classifiers trained on MNIST used an 80-20 train-test split.
	
	\medskip  \noindent \textbf{Training procedures.} Several training procedures have recently been proposed to improve the robustness of neural network classifiers.  Two prominent procedures are the LP-based method in \cite{kolter2017provable} and projected gradient descent (PGD) based method in \cite{madry2017towards}. We refer to these training methods as \texttt{LP-Train} and \texttt{PGD-Train}, respectively. Both procedures take as input a parameter $\epsilon$ that defines the $\ell_{\infty}$ perturbation of the training data points.
	
	
	\medskip \noindent \textbf{Baselines.}  Throughout the experiments, we will often show comparisons to what we call the naive lower and upper bounds.  As has been shown in several previous works \cite{combettes_lipschitz, virmaux2018lipschitz}, trivial lower and upper bounds on the Lipschitz constant of a feed-forward neural network with $\ell$ hidden-layers are given by
	%
	$
	L_{2, \text{ lower}} = \norm{W^{\ell} W^{\ell-1} \cdots W^{0}}_2$ and $L_{2, \text{ upper}} = \prod_{i=0}^{\ell} \norm{W^i}_2$, which we refer to them as naive lower and upper bounds, respectively.
	%
	We are aware of only two methods that bound the Lipschitz constant and can scale to fully-connected networks with more than two hidden layers; these methods are \cite{combettes_lipschitz}, which we will refer to as \texttt{CPLip}, and \cite{virmaux2018lipschitz}, which is called \texttt{SeqLip}. 
	
	%
	
	\begin{figure}[t!]
		\centering
		\begin{subfigure}[t]{.31\textwidth}
			\centering
			\includegraphics[width=\linewidth]{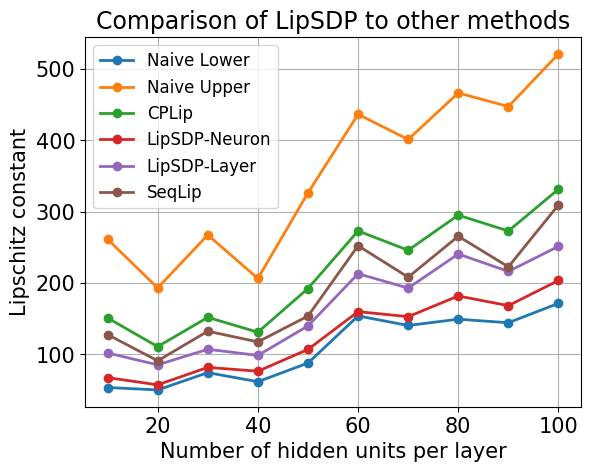}
			\caption{Comparison of Lipschitz bounds found by various methods for five-hidden-layer networks trained on MNIST with the Adam optimizer.  Each network had a test accuracy above 97\%.}
			\label{fig:comp-mnist}
		\end{subfigure} \quad
		\begin{subfigure}[t]{.31\textwidth}
			\centering
			\includegraphics[width=\linewidth]{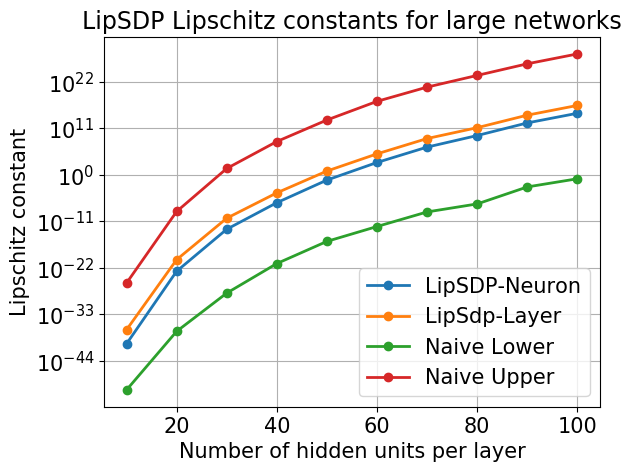}
			\caption{Lipschitz bounds obtained by splitting a 100-layer network into sub-networks.  Each sub-network had six layers, and the weights were generated randomly by sampling from a normal distribution.}
			\label{fig:comp-split}
		\end{subfigure} \quad 
		\caption{Comparison of the accuracy \texttt{LipSDP} methods to other methods that compute the Lipschitz constant and scalability analysis of all three \texttt{SeqLip} methods.}
		\label{fig:comparisons}
	\end{figure}
	
	We compare the Lipschitz bounds obtained by \texttt{LipSDP-Neuron}, \texttt{LipSDP-Layer}, \texttt{CPLip}, and \texttt{SeqLip} in Figure \ref{fig:comp-mnist}.  It is evident from this figure that the bounds from \texttt{LipSDP-Neuron} are tighter than \texttt{CPLip} and \texttt{SeqLip}.  Our results show that the true Lipschitz constants of the  networks shown above are very close to the naive lower bound.
	
	To demonstrate the scalability of the \texttt{LipSDP} formulations, we split a 100-hidden layer neural network into sub-networks with six hidden layers each and computed the Lipschitz bounds using \texttt{LipSDP-Neuron} and \texttt{LipSDP-Layer}.  The results are shown in Figure \ref{fig:comp-split}.  Furthermore, in Tables 1 and 2, we show the computation time for scaling the \texttt{LipSDP} methods in the number of hidden units per layer and in the number of layers.  In particular, the largest network we tested in Table 2 had 50,000 hidden neurons; \texttt{SDPLip-Neuron} took approximately 12 minutes to find a Lipschitz bound, and \texttt{SDPLip-Layer} took approximately 4 minutes.
	
	
	\begin{figure}[t!]
		\centering
		\begin{subfigure}[t]{.48\textwidth}
			\centering
			\includegraphics[width=\linewidth]{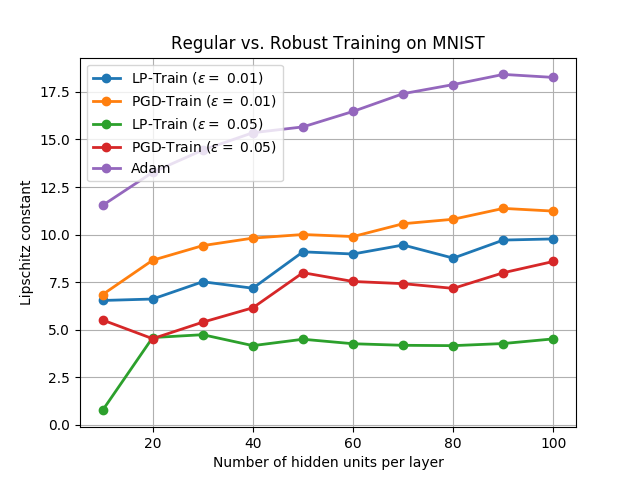}
			\caption{Lipschitz bounds for a one-hidden-layer neural networks trained on the MNIST dataset with the Adam optimizer and \texttt{LP-Train} and \texttt{PGD-Train} for two values of the robustness parameter $\epsilon$.  Each network reached an accuracy of 95\% or higher.}
			\label{fig:rob-vs-reg}
		\end{subfigure} \quad
		\begin{subfigure}[t]{.48\textwidth}
			\centering
			\includegraphics[width=\linewidth]{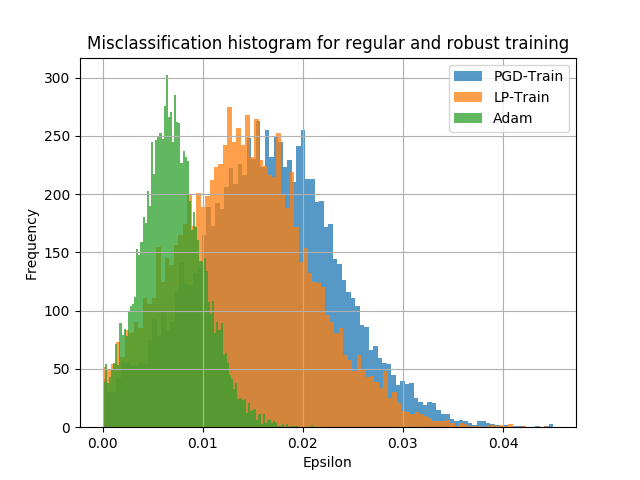}
			\captionof{figure}{Histograms showing the local robustness (in $\ell_{\infty}$ norm) around each correctly-classified test instance from the MNIST dataset.  The neural networks had three hidden layers with 100, 50, 20  neurons, respectively.  All classifiers had a test accuracy of 97\%.}
			\label{fig:histograms}
		\end{subfigure}
		\caption{Analysis of impact of robust training on the Lipschitz constant and the distance to misclassification for networks trained on MNIST}
		\label{fig:robust-training}
	\end{figure}
	
	\medskip \noindent \textbf{Impact of robust training.}   In Figure \ref{fig:robust-training}, we empirically demonstrate that the Lipschitz bound of a neural network is  directly related to the robustness of the corresponding classifier.  This figure shows that \texttt{LP-train} and \texttt{PGD-Train} networks achieve lower Lipschitz bounds than standard training procedures.  Figure \ref{fig:rob-vs-reg} indicates that robust training procedures yield lower Lipschitz constants than networks trained with standard training procedures such as the Adam optimizer.  Figure \ref{fig:histograms} shows the utility of sharply estimating the Lipschitz constant; a lower value of $L_2$ guarantees that a neural network is more locally robust to input perturbations; see Proposition \ref{prop: relation between L and eps} in the Appendix.
	
	\begin{figure}[t!]
		\centering
		\begin{minipage}{.48\textwidth}
			\centering
			\includegraphics[width=\linewidth]{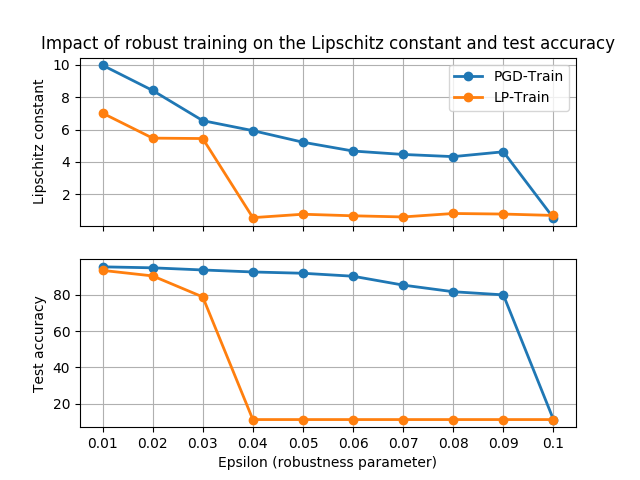}
			\caption{Trade-off between accuracy and Lipschitz constant for different values of the robustness parameter used for \texttt{LP-Train} and \texttt{PGD-Train}.  All networks had one hidden layer with 50 hidden neurons.}
			\label{fig:rob-acc-vs-lip}
		\end{minipage} \quad
		\begin{minipage}{.48\textwidth}
			\centering
			\includegraphics[width=\linewidth]{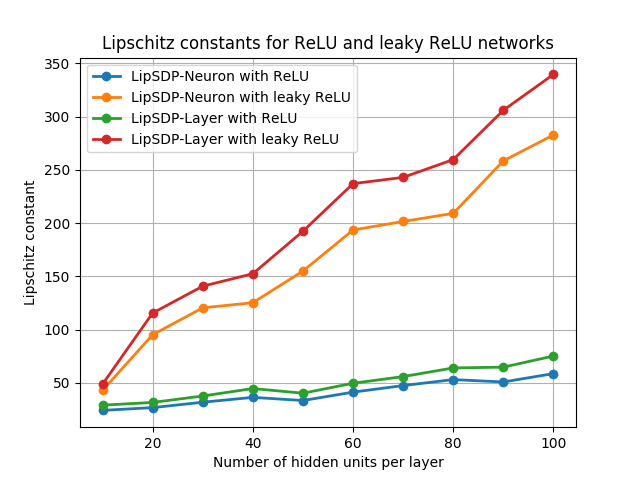}
			\captionof{figure}{Lipschitz constants for topologically identical three-hidden-layer networks with ReLU and leaky ReLU activation functions.  All classifiers were trained until they reached 97\% test accuracy.}
			\label{fig:relu-vs-leaky}
		\end{minipage}
	\end{figure}
	
	
	In the same vein, Figure \ref{fig:rob-acc-vs-lip} shows the impact of varying the robustness parameter $\epsilon$ used in \texttt{LP-Train} and \texttt{PGD-Train} on the test accuracy of networks trained for a fixed number of epochs and the corresponding Lipschitz constants.  In essence, these results quantify how much robustness a fixed classifier can handle before accuracy plummets.   Interestingly, the drops in accuracy as $\epsilon$ increases coincide with corresponding drops in the Lipschitz constant for both \texttt{LP-Train} and \texttt{PGD-Train}.
	
	\medskip \noindent \textbf{Robustness for different activation functions.}  The framework proposed in this work allows us to examine the impact of using different activation functions on the Lipschitz constant of neural networks. We trained two sets of neural networks on the MNIST dataset.  The first set used ReLU activation functions, while the second set used leaky ReLU activations.  Figure \ref{fig:relu-vs-leaky} shows empirically that the networks with the leaky ReLU activation function have larger Lipschitz constants than networks of the same architecture with the ReLU activation function.  
	
	\section{Conclusions and future work}
	In this paper, we proposed a hierarchy of semidefinite  programs to derive tight upper bounds on the Lipschitz constant of feed-forward fully-connected neural networks. Some comments are in order. First, our framework can be directly used to certify convolutional neural networks (CNNs) by unrolling them to a large feed-forward neural network. A future direction is to exploit the special structure of CNNs in the resulting SDP. Second, we only considered one application of Lipschitz bounds in depth (robustness certification). Having an accurate upper bound on the Lipschitz constant can be useful in domains beyond robustness analysis, such as stability analysis of feedback systems with control policies updated by deep reinforcement learning. Furthermore, Lipschitz bounds can be utilized during training as a heuristic to promote out-of-sample generalization \cite{tsuzuku2018lipschitz}. We intend to pursue these applications for future work.
	
	
	\bibliographystyle{plain}
	\bibliography{Refs_nips}
	
	\newpage
	\appendix

	\section{Appendix}
	
	\subsection{Robustness certification of DNN-based classifiers}
	Consider a classifier described by a feed-forward neural network $f \colon \mathbb{R}^{n} \to \mathbb{R}^{k}$, where $n$ is the number of input features and $k$ is the number of classes.  In this context, the function $f$ takes as input an instance or measurement $x$ and returns a $k$-dimensional vector of scores -- one for each class. The classification rule is based on assigning $x$ to the class with the highest score.  That is, we define the classification $C(x):\R^n \rightarrow \{1, \dots, k\}$ to be $C(x)= \mathrm{argmax}_{1 \leq i \leq k} \ f_i(x)$.  Now suppose that $x^{\star}$ is an instance that is classified correctly by the neural network. To evaluate the local robustness of the neural network around $x^\star$, we consider a bounded set $\mathcal{A}(x^\star) = \{x \mid \|x-x^\star\|_2 \leq \epsilon \} \subseteq \mathbb{R}^n$ that represents the set of all possible $\ell_2$-norm perturbations of $x^\star$. Then the classifier is locally robust at $x^\star$ against $\mathcal{A}(x^\star)$ if it assigns all the perturbed inputs to the same class as the unperturbed input, i.e., if
	\begin{align}
		C(x) = C(x^\star) \quad \forall x \in \mathcal{A}(x^\star).
	\end{align}
	In the following proposition, we derive a sufficient condition to guarantee local robustness around $x^\star$ for the perturbation set $\mathcal{A}(x^\star)$.
	\begin{proposition} \label{prop: relation between L and eps}
		
		Consider a neural-network classifier $f\colon \mathbb{R}^n \to \mathbb{R}^k$ with Lipschitz constant $L_2$ in the $\ell_2$-norm. Let $\epsilon>0$ be given and consider the inequality
		\begin{align} \label{eq: local robustness inequality}
			L_2 \leq \frac{1}{\epsilon \sqrt{2}} \min_{1 \leq j \leq k, \ j \neq i^\star } \ |f_j(x^\star)-f_{i^\star}(x^\star)|.
		\end{align}
		Then \eqref{eq: local robustness inequality} implies $C(x):=\mathrm{argmax}_{1 \leq i \leq k} f_i(x) = C(x^\star)$ for all $x\in\mathcal{A}(x^{\star})$.
		
	\end{proposition}
	
	The inequality in \eqref{eq: local robustness inequality} provides us with a simple and computationally efficient test for assessing the point-wise robustness of a neural network. According to \eqref{eq: local robustness inequality}, a more accurate estimation of the Lipschitz constant directly increases the maximum perturbation $\epsilon$ that can be certified for each test example. This makes the framework suitable for model selection, wherein one wishes to select the model that is most robust to adversarial perturbations from a family of proposed classifiers \cite{peck2017lower}.

	\subsection{Proof of Proposition \ref{prop: relation between L and eps}}
	Let $i^\star = \mathrm{argmax}_{1 \leq i \leq k} \ f_i(x^\star)$ be the class of $x^\star$. Define the polytope in the output space of $f$:
	\begin{align*}
		\mathcal{P}_{i^\star} = \{y \mid (e_j-e_{i^\star})^\top y \leq 0 \quad 1 \leq j \leq k, \ j \neq i^\star \}.
	\end{align*}
	which is the set of all outputs whose score is the highest for class $i^\star$; and, of course, $f(x^\star) \in \mathcal{P}_{i^\star}$. The distance of $f(x^\star)$ to the boundary $\partial \mathcal{P}_i$ of the polytope is the minimum distance of $f(x^\star)$ to all edges of the polytope:
	\begin{align*}
		\mathbf{dist}(f(x^\star),\partial \mathcal{P}_{i^\star}) = \inf_{y \in \partial \mathcal{P}_{i^\star}} \|y-f(x^\star)\|_2 = \frac{1} {\sqrt{2}} \min_{1 \leq j \leq k, \ j \neq i^\star } \ |f_j(x^\star)-f_{i^\star}(x^\star)|.
	\end{align*}
	%
	Note that the Lipschitz condition implies $\mathbf{dist}(f(x^\star),f(x))=\|f(x)-f(x^\star)\|_2 \leq L_2 \epsilon$ for all $\|x-x^\star\|_2 \leq \epsilon$. The condition in \eqref{eq: local robustness inequality} then implies
	\begin{align*}
		\mathbf{dist}(f(x^\star),f(x)) \leq \mathbf{dist}(f(x^\star),\partial \mathcal{P}_{i^\star}) \ \text{ for all } \|x-x^\star\|_2 \leq \epsilon,
	\end{align*}
	and for all $1 \leq j \leq k, \ j \neq i^\star$. Therefore, the output of the classification would not change for any $\|x-x^\star\|_2 \leq \epsilon$.
	
	\begin{figure}
		\centering
		\includegraphics[width=0.8\linewidth]{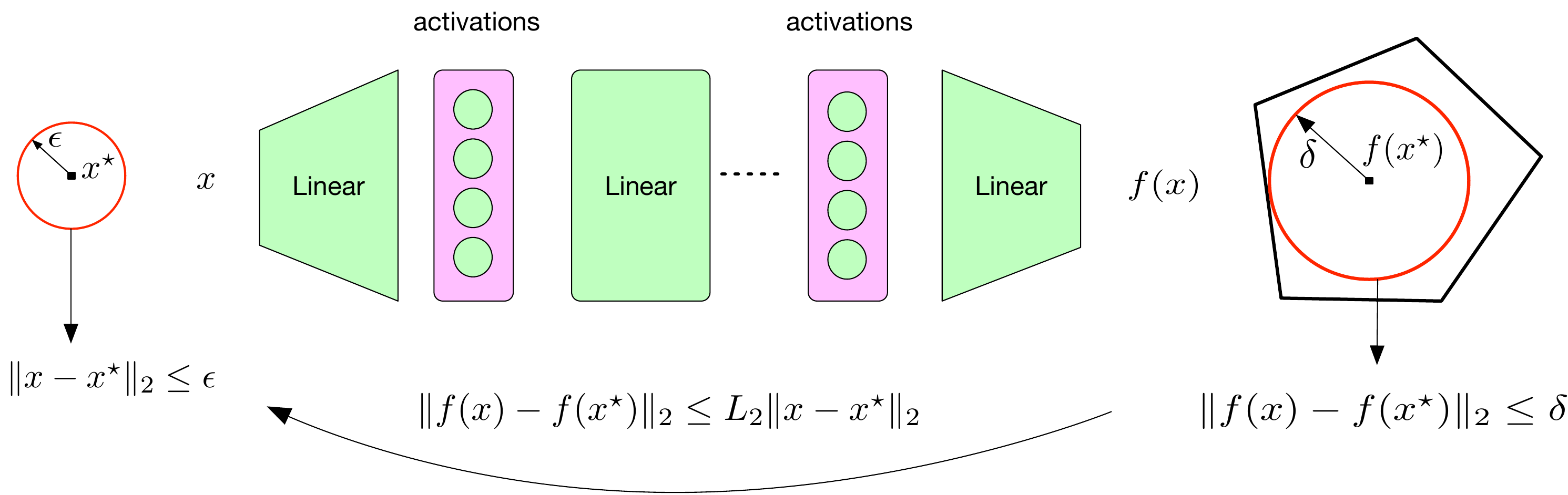}
		\caption{Illustration of local robustness certification using the Lipschitz bound.}
		\label{fig:robust-cert}
	\end{figure}

	\subsection{Proof of Lemma \ref{lem: activation function slope condition vector}}
	%
	\new{ Let $T = \mathrm{diag}(\lambda_{11},\cdots,\lambda_{nn})$ with $\lambda_{ii} \geq 0$ for all $1\leq i \leq n$. By expanding the quadratic form on the right-hand side of  \eqref{eq: activation function slope condition vector}, we can write
		\begin{align}
			&-2\alpha \beta (x\!-\!y)^\top T(x-y) + 2(\alpha+\beta) (x\!-\!y)^\top T(\phi(x)\!-\!\phi(y)) - 2(\phi(x)\!-\!\phi(y))^\top T (\phi(x)\!-\!\phi(y)) \nonumber \\ &= \sum_{i=1}^{n} \lambda_{ii} \left(-2\alpha \beta (x_i-y_i)^2+2(\alpha+\beta)(x_i-y_i)(\varphi(x_i)-\varphi(y_i))-2(\varphi(x_i)-\varphi(y_i))^2\right) \nonumber \\ &= \sum_{i=1}^{n} \lambda_{ii} \begin{bmatrix}
				x_i-y_i \\ \varphi(x_i)-\varphi(y_i)
			\end{bmatrix}^\top \begin{bmatrix}
				-2\alpha \beta & \alpha+\beta \\ \alpha+\beta & -2
			\end{bmatrix}\begin{bmatrix}
				x_i-y_i \\ \varphi(x_i)-\varphi(y_i)
			\end{bmatrix} \geq 0, \nonumber 
		\end{align}
		where the last inequality follows from \eqref{eq: slope restricted qc}. 
	}
	
	\subsection{Proof of Theorem \ref{thm: Lipschitz one layer}} 	
	Define $x^1 = \phi(W^0 x + b^0) \in \mathbb{R}^{n}$ and $y^1 = \phi(W^0 y + b^0) \in \mathbb{R}^{n}$ for two arbitrary inputs $x, y \in \mathbb{R}^{n_0}$. Using Lemma \ref{lem: activation function slope condition vector}, we can write the quadratic inequality
	\begin{align*} 
		0 \leq \begin{bmatrix}
			(W^0x+b^0)-(W^0 y+b^0) \\ x^1-y^1
		\end{bmatrix}^\top \begin{bmatrix}
			-2\alpha \beta T & (\alpha+\beta)T \\ (\alpha+\beta)T & -2T
		\end{bmatrix} \begin{bmatrix}
			(W^0x+b^0)-(W^0 y+b^0) \\ x^1-y^1
		\end{bmatrix},
	\end{align*}
	where $T \in \mathcal{T}_n$ and $\mathcal{T}_n$ is defined as in \eqref{lem: activation function slope condition vector 1}. The preceding inequality can be simplified to
	\begin{align} 
		0 \leq \begin{bmatrix}
			x\!-\!y \\ x^1\!-\!y^1
		\end{bmatrix}^\top 	\begin{bmatrix}
			-2\alpha \beta {W^0}^\top T W^0  & (\alpha+\beta) {W^0}^\top T  \\ (\alpha+\beta) TW^0 & -2T
		\end{bmatrix}
		\begin{bmatrix}
			x\!-\!y \\ x^1\!-\!y^1
		\end{bmatrix}.
	\end{align}
	By left and right multiplying $M(\rho,T)$ in \eqref{thm: Lipschitz one layer 2} by $\begin{bmatrix}(x\!-\!y)^\top & (x^1\!-\!y^1)^\top \end{bmatrix}$ and $\begin{bmatrix}(x\!-\!y)^\top & (x^1\!-\!y^1)^\top \end{bmatrix}^\top$, respectively, and rearranging terms, we obtain
	\begin{align} \label{thm: Lipschitz one layer 6}
		\begin{bmatrix}
			x\!-\!y \\ x^1\!-\!y^1
		\end{bmatrix}^\top 	\begin{bmatrix}
			-2\alpha \beta {W^0}^\top T W^0  & (\alpha+\beta) {W^0}^\top T  \\ (\alpha+\beta) TW^0 & -2T
		\end{bmatrix} \begin{bmatrix}
			x\!-\!y \\ x^1\!-\!y^1
		\end{bmatrix} \\ \leq \begin{bmatrix}
			x\!-\!y \\ x^1\!-\!y^1
		\end{bmatrix}^\top 	\begin{bmatrix}
			L_2^2 I_{n_0}  & 0  \\ 0 & -{W^1}^\top W^1
		\end{bmatrix} \begin{bmatrix}
			x\!-\!y \\ x^1\!-\!y^1
		\end{bmatrix}. \notag
	\end{align}
	By adding both sides of the preceding inequalities, we obtain
	\begin{align*}
		0 \leq \begin{bmatrix}
			x\!-\!y \\ x^1\!-\!y^1
		\end{bmatrix}^\top 	\begin{bmatrix}
			L_2^2 I_{n_0}  & 0  \\ 0 & -{W^1}^\top W^1
		\end{bmatrix} \begin{bmatrix}
			x\!-\!y \\ x^1\!-\!y^1
		\end{bmatrix},
	\end{align*}
	or, equivalently,
	\begin{align*}
		(x^1-y^1)^\top {W^1}^\top W^1 (x^1-y^1) \leq L_2^2 (x-y)^\top (x-y).
	\end{align*}
	Finally, note that by definition of $x^1$ and $y^1$, we have $f(x)=W^1x^1+b^1$ and $f(y)=W^1 y^1+b^1$. Therefore, the preceding inequality implies
	\begin{align}
		\|f(x)-f(y)\|_2 \leq L_2 \|x-y\|_2 \quad \text{for all } x,y \in \mathbb{R}^{n_0}.
	\end{align}
	\subsection{Proof of Theorem \ref{thm:multi-layer-thm}} \label{thm:multi-layer-thm proof}
	For two arbitrary inputs $x^0,y^0 \in \mathbb{R}^{n_0}$, define $\bx = [{x^0}^\top \cdots {x^{\ell}}^\top]^\top \quad \text{ and }  \quad \by = [{y^0}^\top \cdots {y^{\ell}}^\top]^\top$
	Using the compact notation in \eqref{eq:multi-layer-eqs}, we can write
	\begin{align*}
		B\bx &= \phi(A\mathbf{x} + b) \quad \text{ and }  \quad B\by = \phi(A\mathbf{y} + b).
	\end{align*}
	Multiply both sides of the first matrix in \eqref{thm:multi-layer-thm 2} by $(\bx-\by)^\top$ and $(\bx-\by)$, respectively and use the preceding identities to obtain
	\begin{align} \label{thm:multi-layer-thm 5}
		&(\bx-\by)^\top \begin{bmatrix}
			A \\ B
		\end{bmatrix}^\top \begin{bmatrix}
			-2\alpha \beta T & (\alpha+\beta)T \\ (\alpha+\beta)T & -2T
		\end{bmatrix} \begin{bmatrix}
			A \\ B
		\end{bmatrix}(\bx-\by)  \\
		&=\begin{bmatrix}
			A \bx-A\by \\ B \bx - B \by
		\end{bmatrix}^\top \begin{bmatrix}
			-2\alpha \beta T & (\alpha+\beta)T \\ (\alpha+\beta)T & -2T
		\end{bmatrix} \begin{bmatrix}
			A\bx-A\by \\ B\bx-B\by
		\end{bmatrix} \notag
		\\
		&=\begin{bmatrix}
			A \bx-A\by \\ \phi(A\bx+b)-\phi(A\by+b)
		\end{bmatrix}^\top \begin{bmatrix}
			-2\alpha \beta T & (\alpha+\beta)T \\ (\alpha+\beta)T & -2T
		\end{bmatrix} \begin{bmatrix}
			A\bx-A\by \\ \phi(A\bx+b)-\phi(A\by+b)
		\end{bmatrix} \geq 0, \notag
	\end{align}
	where the last inequality follows from Lemma \ref{lem: activation function slope condition vector}. On the other hand, by multiplying both sides of the second matrix in \eqref{thm:multi-layer-thm 2} by $(\bx-\by)^\top$ and $(\bx-\by)$, respectively, we can write
	\begin{align} \label{thm:multi-layer-thm 6}
		(\bx-\by)^\top M (\bx-\by) = \|f(x)-f(y)\|_2^2 - L_2^2 \|x-y\|_2^2,
	\end{align}
	where we have used the fact that $f(x)=W^\ell x^\ell + b^\ell$ and $f(y)=W^\ell y^\ell+b^\ell$. By adding both sides of \eqref{thm:multi-layer-thm 5} and \eqref{thm:multi-layer-thm 6}, we get
	\begin{align} \label{thm:multi-layer-thm 7}
		(\bx-\by)^\top\left( \begin{bmatrix}
			A \\ B
		\end{bmatrix}^\top \begin{bmatrix}
			-2\alpha \beta T & (\alpha+\beta)T \\ (\alpha+\beta)T & -2T
		\end{bmatrix} \begin{bmatrix}
			A \\ B
		\end{bmatrix}+M\right)(\bx-\by) \geq  \|f(x)-f(y)\|_2^2 - L_2^2 \|x-y\|_2^2.
	\end{align}
	When the LMI in \eqref{thm:multi-layer-thm 2} holds, the left-hand side of \eqref{thm:multi-layer-thm 7} is non-positive, implying that the right-hand side is non-positive.

	
	\subsection{Bounding the output set of a neural network classifier}
	
	\begin{figure}[t!]
		\centering
		\includegraphics[width=0.7\linewidth]{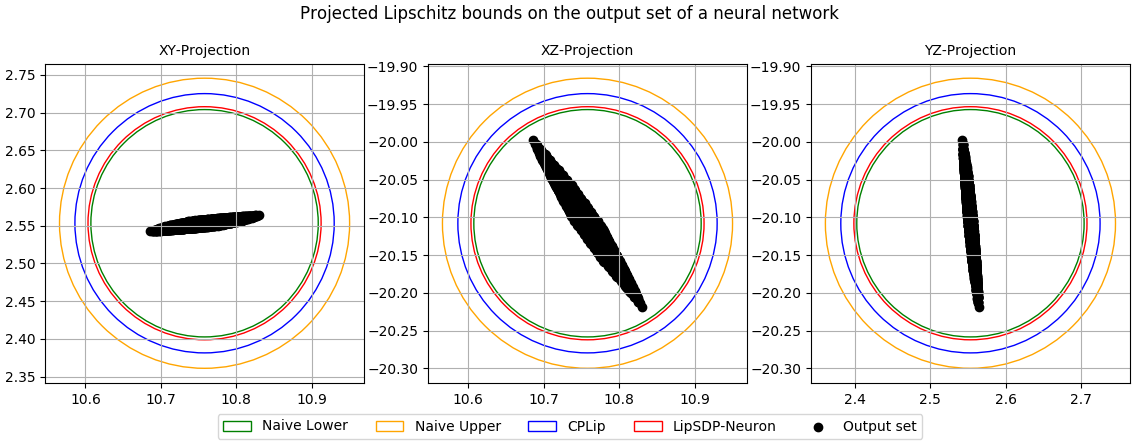}
		\caption{Bounds on the image of an $\ell_2$ $0.01$-ball under a neural network trained on the Iris dataset.  The network was trained using the Adam optimizer and reached a test accuracy of 99\%.}
		\label{fig:iris}
	\end{figure}
	
	As we have shown, accurate estimation of the Lipschitz constant can provide tighter bounds on adversarial examples drawn from a perturbation set around a nominal point $\bar{x}$.  Figure \ref{fig:iris} shows these bounds for a neural network trained on the Iris dataset.  Because this dataset has three classes, we project the output sets onto the coordinate axes.  This figure clearly shows that our bound is quite close to the naive lower bound.  Indeed, the naive upper bound and the constant computed by \texttt{CPLip} are considerably looser.
	
	\subsection{Further analysis of scalability} 
	
	\begin{figure}[t!]
		\centering
		\includegraphics[width=0.45\linewidth]{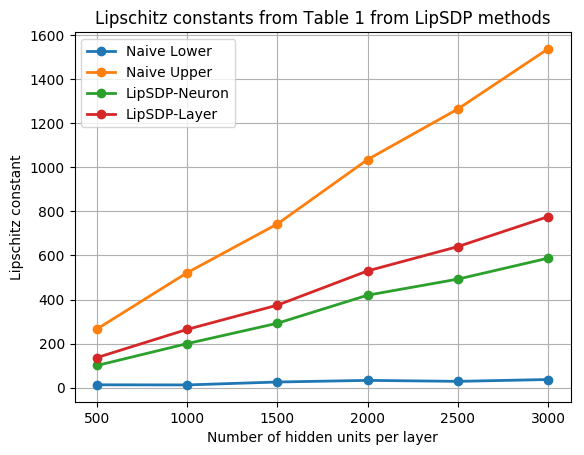}
		\caption{Lipschitz bounds for networks in Table 1.}
		\label{fig:table1}
	\end{figure}
	
	Figure \ref{fig:table1} shows the Lipschitz bounds found for the networks used in Table 1.
	\new{
		\subsection{LipSDP and repeated nonlinearities} \label{appendix: LipSDP and repeated nonlinearities}
		The incremental quadratic constraint in Lemma \ref{lem: activation function slope condition vector} is valid for functions of the form $\phi(x)= [\varphi_1(x_1) \cdots \varphi_n(x_n)]^\top$ as along as all $\varphi_i$'s are slope-restricted on $[\alpha,\beta]$. Therefore, this quadratic constraint does not exploit the fact that the same activation functions are used. In this subsection, we derive more refined quadratic constraints that exploit this additional structure when one of the points in the definition of Lipschitz continuity is fixed. Explicity, for a given $y \in \mathbb{R}^{n_0}$, our goal is to compute the smallest $\rho$ (depending on $y$) such that 
		\begin{align} \label{eq: lip constant fixed point}
			\|f(x)-f(y)\|_2 \leq  \sqrt{\rho} \|x-y\|_2 \quad \forall x \in \mathbb{R}^{n_0}
		\end{align}
		This inequality can be equivalently written as 
		\begin{align}
			\begin{bmatrix}
				x \\  f(x) \\ 1
			\end{bmatrix}^\top
			\underbrace{\begin{bmatrix}
					-\rho I_{n_0}	& 0 & \rho y \\ 0 & I_{n_f} & -f(y) \\ \rho y^\top & -f(y)^\top & f(y)^\top f(y)-\rho  y^\top y
			\end{bmatrix}}_{R(\rho)}
			\begin{bmatrix}
				x \\  f(x) \\ 1
			\end{bmatrix} \leq 0 	\quad \forall x \in \mathbb{R}^{n_0}	
		\end{align}
		To certify the quadratic inequality above, we make use of the following definition from \cite{fazlyab2019safety}.
		\begin{definition} \label{eq: QC def 0}
			Let $\phi \colon \mathbb{R}^{n} \to \mathbb{R}^{n}$ and suppose $\mathcal{Q} \subset \mathbb{S}^{2n+1}$ is the set of all symmetric and indefinite matrices $Q$ such that the inequality
			\begin{align} \label{eq: QC def}
				\begin{bmatrix}
					x \\ \phi(x) \\ 1
				\end{bmatrix}^\top Q \begin{bmatrix}
					x \\ \phi(x) \\ 1
				\end{bmatrix} \geq 0,
			\end{align}
			holds for all $x \in \mathcal{X} \subseteq \mathbb{R}^{n}$. Then we say $\phi$ satisfies the quadratic constraint defined by $\mathcal{Q}$ on $\mathcal{X}$.
		\end{definition}
		Note that Quadratic Constraints as defined above are more general than Incremental Quadratic Constraints, and would allow us to capture more information about $\phi$. For example, for any pair $(x_i,x_j) \ i \neq j$, since $\varphi(x_i)$ and $\varphi(x_j)$ are obtained from the same nonlinearity, we can write the incremental quadratic constraint
		\begin{align*}
			\begin{bmatrix}
				x_i - x_j \\ \varphi(x_i)-\varphi(x_j)
			\end{bmatrix}^\top \begin{bmatrix}
				-2\alpha \beta & \alpha +\beta \\ \alpha+\beta & -2
			\end{bmatrix} \begin{bmatrix}
				x_i - x_j \\ \varphi(x_i)-\varphi(x_j)
			\end{bmatrix} \geq 0 \quad  1 \leq i <j \leq n,
		\end{align*}
		This inequality can be equivalently represented as 
		\begin{align}
			\begin{bmatrix}
				x_i \\ x_j \\ \varphi(x_i) \\ \varphi(x_j) \\1
			\end{bmatrix}^\top \begin{bmatrix}
				-2 \alpha \beta & 2 \alpha \beta & \alpha+\beta & -(\alpha+\beta) & 0
				\\ 2 \alpha \beta & -2 \alpha \beta & -(\alpha+\beta) & \alpha+\beta & 0\\
				\alpha+\beta & -(\alpha+\beta) & -2 & 2 & 0 \\
				-(\alpha+\beta) & \alpha+\beta & 2 & -2 & 0 \\
				0 & 0 & 0 & 0 & 0
			\end{bmatrix} \begin{bmatrix}
				x_i \\ x_j \\ \varphi(x_i) \\ \varphi(x_j) \\1
			\end{bmatrix} \geq 0,
		\end{align}		
		This quadratic constraints couples neuron $i$ and $j$, $i \neq j$, resulting in $O(n^2)$ quadratic constraints for an activation layer with $n$ neurons. In the following lemma, we characterize $\phi(x) = [\varphi(x_1) \cdots \varphi(x_n)]^\top$ with all possible incremental quadratic constraints.
		\begin{lemma}[Lemma 2 of \cite{fazlyab2019safety}] \label{lem: activation function slope condition repeated}
			Suppose $\varphi \colon \mathbb{R} \to \mathbb{R}$ is slope-restricted on $[\alpha,\beta]$. Define the set
			\begin{align} \label{lem: activation function slope condition vector 2}
				\mathcal{S}_{n} = \left\{S \in \mathbb{S}^{n} \mid  S \!=\! \!\sum_{1\leq i<j \leq n} \lambda_{ij}(e_i\!-\!e_j)(e_i\!-\!e_j)^\top, \ \lambda_{ij} \geq 0 \right\}.
			\end{align}
			where $e_i$ is the $i$-th standard basis vector in $\mathbb{R}^{n}$. Then for any $S \in \mathcal{S}_{n}$ the vector-valued function $\phi:\R^n\rightarrow\R^n$ defined by $\phi(x) = [\varphi(x_1) \cdots \varphi(x_n)]^\top$ satisfies
			\begin{align} \label{eq: activation function slope condition repeated}
				\begin{bmatrix}
					x \\ \phi(x) \\ 1
				\end{bmatrix}^\top \begin{bmatrix}
					-2\alpha \beta S & (\alpha+\beta)S  & 0 \\ (\alpha+\beta)S & -2S & 0 \\ 0 & 0 & 0
				\end{bmatrix}  \begin{bmatrix}
					x \\ \phi(x) \\ 1
				\end{bmatrix} \geq 0 \ \text{ for all } x,y \in \mathbb{R}^n.
			\end{align}
		\end{lemma}
		In the above Lemma, we are only capturing the quadratic constraints implied by slope restriction that are imposed between any pair of neurons. However, depending on what activation function is being considered, more quadratic constraints can be included. For a detailed account of Quadratic Constraints, see \cite{fazlyab2019safety}.
		
		\paragraph{LipSDP with repeated nonlinearities.} Consider the compact representation \eqref{eq:multi-layer-eqs} of a multi-layer network $f$ defined by the recursion \eqref{eq: DNN model 0}. Suppose $\phi$ satisfies the quadratic constraint defined by $\mathcal{Q}$ on $\mathbb{R}^n$. Utilizing Definition \ref{eq: QC def 0}, we can write the following quadratic constraints for any $Q \in \mathcal{Q}$,
		\begin{align} \label{eq: M_mid}
			\begin{bmatrix}
				A \bx + b \\ B \bx \\ 1
			\end{bmatrix}^\top Q   \begin{bmatrix}
				A \bx + b \\ B \bx \\ 1
			\end{bmatrix}  = \begin{bmatrix}
				\bx \\  1
			\end{bmatrix}^\top \begin{bmatrix}
				A &  b\\ B & 0 \\ 0 & 1
			\end{bmatrix}^\top Q \begin{bmatrix}
				A &  b\\ B & 0 \\ 0 & 1
			\end{bmatrix}\begin{bmatrix}
				\bx \\ 1
			\end{bmatrix} \geq 0, \quad \forall \mathbf{x}
		\end{align}
		Now consider the following LMI,
		\begin{align} \label{thm:multi-layer-thm 3}
			N(\rho,T) \!:=\!  \begin{bmatrix}
				A &  b\\ B & 0 \\ 0 & 1
			\end{bmatrix}^\top Q \begin{bmatrix}
				A &  b\\ B & 0 \\ 0 & 1
			\end{bmatrix} \!+\! \begin{bmatrix}
				I_{n_0} & 0 & 0 \\
				0 & 0 & 0 \\
				\vdots & \vdots & \vdots \\
				0 & {W^{\ell}}^\top & 0 \\ 0 &  {b^{\ell}}^\top & 1
			\end{bmatrix} R(\rho) \begin{bmatrix}
				I_{n_0} & 0 & 0 \\
				0 & 0 & 0 \\
				\vdots & \vdots & \vdots \\
				0 & {W^{\ell}}^\top & 0 \\ 0 &  {b^{\ell}}^\top & 1
			\end{bmatrix}^\top \preceq 0,
		\end{align}
		where $R(\rho)$ is defined in \eqref{eq: lip constant fixed point}. 
		By left- and right- multiplying both sides by $\begin{bmatrix} \bx^\top  &1\end{bmatrix}$ and $\begin{bmatrix} \bx^\top & 1\end{bmatrix}^\top$, respectively, we obtain
		\begin{align*}
			\begin{bmatrix}
				\bx \\  1
			\end{bmatrix}^\top  
			\left(
			\begin{bmatrix}
				A &  b\\ B & 0 \\ 0 & 1
			\end{bmatrix}^\top Q \begin{bmatrix}
				A &  b\\ B & 0 \\ 0 & 1
			\end{bmatrix} \!+\! \begin{bmatrix}
				I_{n_0} & 0 & 0 \\
				0 & 0 & 0 \\
				\vdots & \vdots & \vdots \\
				0 & {W^{\ell}}^\top & 0 \\ 0 &  {b^{\ell}}^\top & 1
			\end{bmatrix} R(\rho) \begin{bmatrix}
				I_{n_0} & 0 & 0 \\
				0 & 0 & 0 \\
				\vdots & \vdots & \vdots \\
				0 & {W^{\ell}}^\top & 0 \\ 0 &  {b^{\ell}}^\top & 1
			\end{bmatrix}^\top
			\right)
			\begin{bmatrix}
				\bx \\ 1
			\end{bmatrix} \leq 0.
		\end{align*}
		The first term is non-negative by \eqref{eq: M_mid}, implying that the second term is non-positive. Therefore,
		\begin{align*}
			&\begin{bmatrix}
				\bx  \\ 1
			\end{bmatrix}^\top\begin{bmatrix}
				I_{n_0} & 0 & 0 \\
				0 & 0 & 0 \\
				\vdots & \vdots & \vdots \\
				0 & {W^{\ell}}^\top & 0 \\ 0 &  {b^{\ell}}^\top & 1
			\end{bmatrix} R(\rho) \begin{bmatrix}
				I_{n_0} & 0 & 0 \\
				0 & 0 & 0 \\
				\vdots & \vdots & \vdots \\
				0 & {W^{\ell}}^\top & 0 \\ 0 &  {b^{\ell}}^\top & 1
			\end{bmatrix}^\top \begin{bmatrix}
				\bx  \\1
			\end{bmatrix} \! 
			%
			= \begin{bmatrix}
				x^0 \\ f(x^0) \\ 1
			\end{bmatrix}^\top 
			R(\rho)
			\begin{bmatrix}
				x^0 \\ f(x^0) \\ 1
			\end{bmatrix} \leq 0
		\end{align*}
		Therefore, the feasibility of the LMI in \eqref{thm:multi-layer-thm 3} implies the inequality in \eqref{eq: lip constant fixed point}.  We remark that the LMI in \eqref{thm:multi-layer-thm 3} has now $O(n^2)$ decision variables. For future work, we will invesitigate the effectiveness of these additional decision variables on the computed Lipschitz bound. 
	}

\end{document}